\documentclass[journal]{IEEEtran}
\usepackage{cite}
\usepackage{microtype}
\usepackage{amsmath,amssymb,amsfonts}
\usepackage{graphicx}
\usepackage{booktabs,multirow,array}
\usepackage{mathrsfs}
\usepackage{textcomp}
\usepackage{url}
\usepackage[hidelinks]{hyperref}
\graphicspath{{figures/}}
\begin{document}
\title{Anchor and Adapt: Asymmetric Prompt Adaptation for Few-Shot Industrial Anomaly Detection}
\author{Mengyang Zhao, Teng Fu, Haiyang Yu, Ke Niu, Bin Li, and Xiangyang Xue%
\thanks{The authors are with the Shanghai Key Lab of Intelligent Information Processing and College of Computer Science and Artificial Intelligence, Fudan University, Shanghai 200433, China (e-mail: myzhao20@fudan.edu.cn; tfu23@m.fudan.edu.cn; hyyu20@fudan.edu.cn; kniu223@m.fudan.edu.cn; libin@fudan.edu.cn; xyxue@fudan.edu.cn).}%
\thanks{Corresponding author: Xiangyang Xue.}}
\markboth{IEEE Transactions on Circuits and Systems for Video Technology}%
{Zhao \MakeLowercase{\textit{et al.}}: Anchor and Adapt}
\maketitle

\begin{abstract}
In few-shot industrial anomaly detection, the few normal target images provide no direct defect supervision, making anomaly prompts difficult to learn from these samples alone. Some vision-language methods therefore use manually specified descriptions to supply explicit anomaly semantics. However, constructing these descriptions requires product-specific effort, and their effectiveness depends on prompt selection. We propose Anchor and Adapt, a two-stage prompt learning framework that separates the acquisition of anomaly semantics from adaptation to target normal appearance. Stage I learns transferable normal and abnormal anchors from annotated auxiliary data. Stage II keeps these anchors fixed and adapts an additional normal branch using the few target normal samples. The inherited and adapted normal branches jointly characterize target normality, with text-anchor regularization encouraging consistency with the generic normal prior and separation from the abnormal anchors. This design retains learned anomaly knowledge while reducing dependence on category-specific anomaly templates, without requiring synthetic anomaly generation. Cross-dataset experiments between MVTec-AD and VisA under 1-, 2-, and 4-shot settings demonstrate competitive detection and localization performance. Controlled ablations assess the roles of transferred anchors, asymmetric adaptation, dual-normal representations, and anchor regularization.

\end{abstract}


\begin{figure*}[t]
  \centering
  \includegraphics[width=\textwidth]{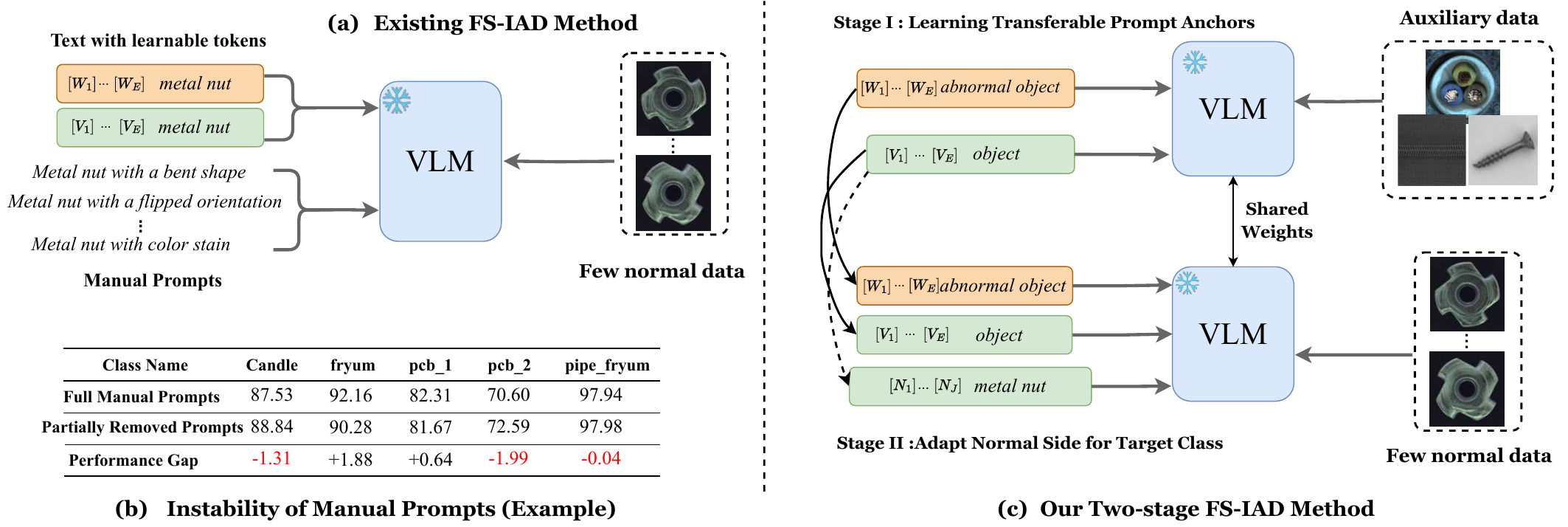}

\caption{
(a) Template-based prompt learning, exemplified by PromptAD, combines manually specified normal/abnormal descriptions with learnable tokens.
(b) Removing two manual prompts per category from PromptAD changes performance in different directions across categories, illustrating sensitivity to prompt selection and motivating reduced reliance on category-specific anomaly templates.
(c) Anchor and Adapt first learns transferable normal/abnormal anchors from auxiliary anomaly supervision, then retains these references while adapting an additional normal branch to the target product using its few normal samples.
$\mathbf{V}$, $\mathbf{W}$, and $\mathbf{N}$ denote learnable prompt tokens.
}

  \label{fig:1}
\end{figure*}

\section{Introduction}

Visual anomaly detection aims to identify images and local regions that deviate from normal visual patterns. It plays an important role in automated industrial quality inspection, enabling the detection of defective products and the localization of defects~\cite{DBLP:journals/tcsv/WangZHWD25,DBLP:journals/tcsv/ChenLGLZ25,filo++,icml-iad1}. Few-shot industrial anomaly detection (FS-IAD) addresses deployment to new product categories where only a few normal samples are available for model adaptation~\cite{inctrl,anomalygpt,metausa,oneforall,filo++}. For vision-language model based approaches ~\cite{promptad,oneforall}, this data constraint places particular importance on how normal and abnormal semantic references are constructed.

Vision-language models (VLMs), such as CLIP~\cite{clip}, support FS-IAD by comparing visual features with textual descriptions of normal and abnormal states. The few normal target samples provide no direct defect supervision, making anomaly prompts difficult to learn from these samples alone. To supply explicit anomaly semantics under this limited supervision, some methods incorporate manually specified descriptions of abnormal states~\cite{winclip,promptad,oneforall,finegrainedad2026}. For example, WinCLIP~\cite{winclip} constructs manually specified prompt ensembles, while PromptAD~\cite{promptad} combines manual descriptions with learnable context tokens optimized from normal target samples, as illustrated in Figure~\ref{fig:1}-a.

However, specifying anomaly descriptions for each product requires repeated design and selection, and adding descriptions does not necessarily improve detection. Figure~\ref{fig:1}-b illustrates this issue: reducing PromptAD's manual prompt set improves performance for some products and degrades it for others. This observation motivates reducing dependence on category-specific anomaly descriptions while retaining useful semantic guidance. The challenge is to obtain useful anomaly semantics while adapting to a new product's normal appearance from only a few normal samples.

The supervision needed to establish anomaly semantics differs from that available for adapting to a new product. Annotated auxiliary data provide direct defect supervision and support learning transferable normal and abnormal semantic references~\cite{anomalyclip,vcp-clip,adaclip}. The few normal target images, in turn, provide evidence of the new product's normal appearance. These complementary sources of supervision motivate a two-stage approach: first acquire transferable semantic references from auxiliary data, then adapt to target normality while retaining those references.

We therefore propose \textbf{Anchor and Adapt}, a two-stage prompt learning framework that assigns complementary roles to auxiliary anomaly supervision and target normal supervision. Stage I learns generic normal and abnormal anchors from annotated auxiliary data. Stage II uses the few target normal samples to specialize the normal representation. Because these samples directly characterize normal appearance, we adopt asymmetric adaptation: the learned anchors remain fixed, while an additional class-specific normal branch is adapted to the target product.

This additional branch complements the inherited generic normal representation. The fixed generic branch retains the transferable normal reference, while the class-specific branch captures the target product's normal appearance; together, they form a dual-normal representation. Text-anchor regularization guides the adapted prompts toward consistency with the generic normal prior and separation from the abnormal anchors. Low-rank residual updates and additional context tokens provide the capacity for this adaptation. The resulting framework combines learned anomaly semantics with target-specific normal evidence without requiring category-specific anomaly templates or synthetic anomaly generation.

We evaluate cross-dataset transfer in both directions between MVTec-AD and VisA under 1-, 2-, and 4-shot target adaptation. Additional experiments on MPDD examine transfer to metal-part inspection. Comparative results and controlled ablations assess the framework and the roles of transferred anchors and target adaptation. Our contributions are as follows:
\begin{itemize}
\item We propose a two-stage prompt learning framework that uses complementary auxiliary anomaly supervision and target normal supervision to acquire transferable semantic anchors and adapt product-specific normality, reducing reliance on manually specified category-specific anomaly descriptions.
\item We develop asymmetric target adaptation that preserves the transferred normal and abnormal anchors while learning an additional normal branch. A dual-normal representation combines inherited and target-specific normal evidence, with anchor regularization guiding adaptation relative to both sets of fixed references.
\item Cross-dataset experiments between MVTec-AD and VisA evaluate detection and localization performance. Controlled ablations examine the roles of learned anchors and the target-adaptation mechanisms, including the gains beyond normal-memory retrieval.
\end{itemize}

The remainder of this paper is organized as follows. Section~\ref{sec:related_work} reviews related work, and Section~\ref{sec:preliminary} introduces the preliminaries. Section~\ref{sec:method} presents the proposed framework, Section~\ref{sec:experiments} reports the experiments, and Section~\ref{sec:conclusion} concludes the paper.

\section{Related Work}
\label{sec:related_work}
\subsection{Vision-Language Models}

Vision-language models such as CLIP~\cite{clip} and ALIGN~\cite{align} learn a shared embedding space for images and text. CLIP aligns visual and textual features through contrastive learning on 400 million image--text pairs. With textual prompts such as ``a photo of a dog'', it performs zero-shot classification by matching image and text features using cosine similarity. Selecting effective manual prompts requires task-specific tuning. CoOp~\cite{CoOp} replaces context words with learnable vectors optimized using a few labeled images. CoCoOp~\cite{CoCoOp} extends this approach with a lightweight network that generates an input-conditional token for each image. These image-conditioned prompts improve generalization to class shift compared with the static prompts of CoOp. Related vision-language approaches have also been applied to image segmentation~\cite{seg-clip}, video understanding~\cite{videoclip}, and Chinese text recognition through image--structure alignment~\cite{yu2023ccrclip}. More recently, task-specific VLM adaptation has enabled interactive person retrieval in ChatReID~\cite{niu2025chatreid} and unified medical image analysis in UMIT~\cite{yu2025umit}. For tracking, FDTracker~\cite{fu2025fdtracker} uses language-guided appearance features for object association, while OmniPT~\cite{fu2026omnipt} integrates pedestrian tracking with interactive semantic understanding.

\begin{figure*}[t]
  \centering
  \includegraphics[width=\textwidth]{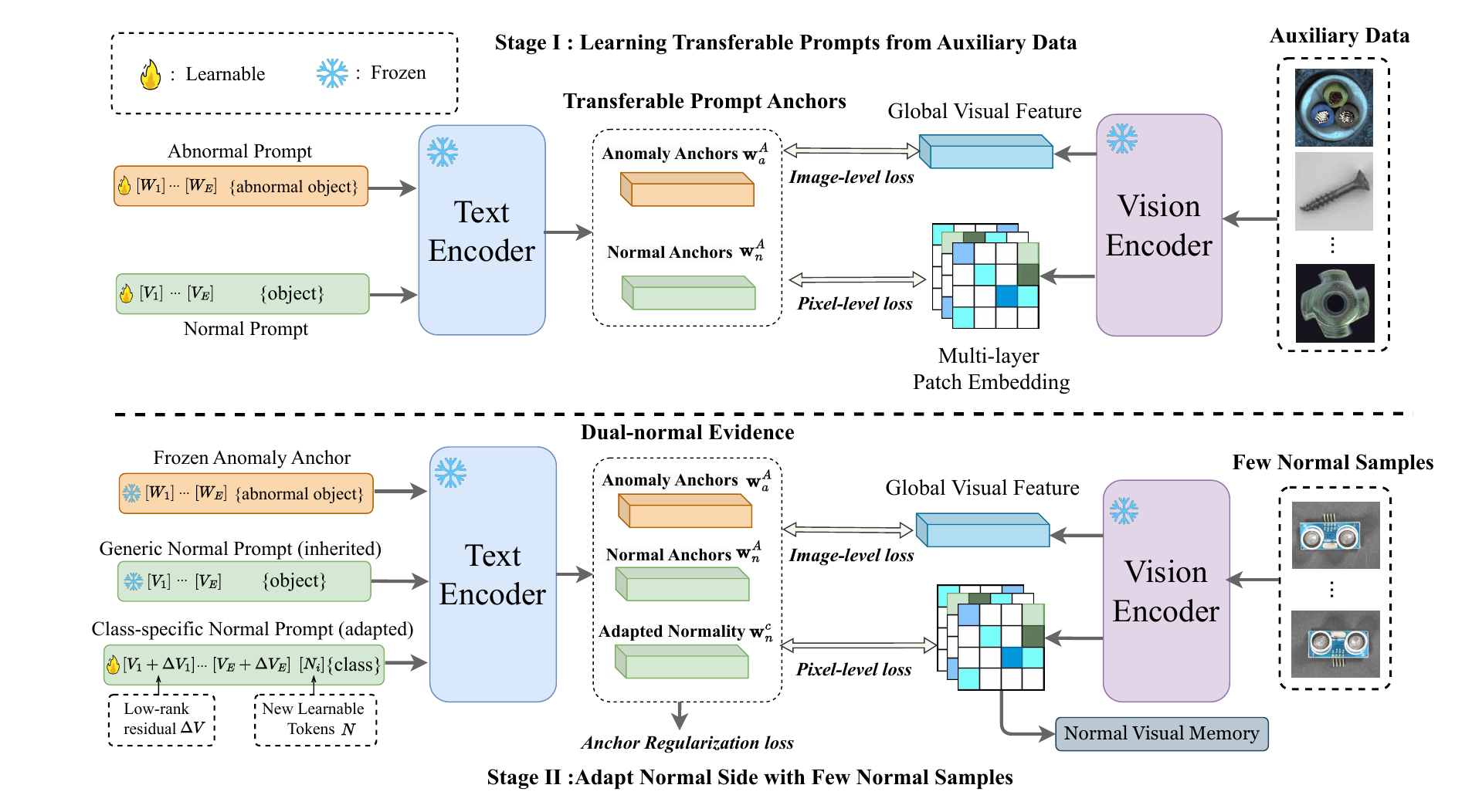}
\caption{
Overview of the two-stage \textbf{Anchor and Adapt} framework.
Stage I pretrains generic normal/abnormal prompt anchors using auxiliary anomaly supervision.
Stage II keeps both sets of anchors fixed and learns a class-specific normal branch from the few target normal samples. The inherited generic branch and adapted class-specific branch jointly provide normal evidence, with text-anchor regularization linking the adapted prompts to the fixed references.
Icons indicate learnable and frozen parameters.
}

  \label{fig:framework}
\end{figure*}

\subsection{Industrial Anomaly Detection}
Industrial anomaly detection aims to detect and localize defects in products. Unsupervised approaches learn normality from training images or use synthesized anomalies to obtain additional supervision~\cite{cutpaste,2023_iccv,Tien_2023_CVPR}. In FS-IAD, the limited number of normal target images places greater emphasis on knowledge transferred from pretrained models or auxiliary datasets. WinCLIP~\cite{winclip} uses normal and abnormal text ensembles. PromptAD~\cite{promptad} learns normal and abnormal prompts from normal target images through semantic concatenation and margin constraints, using manually specified anomaly descriptions to guide prompt learning. FineGrainedAD~\cite{finegrainedad2026} introduces fine-grained visual-language alignment. AnoPLe~\cite{anople2026} combines multimodal prompt learning with simulated anomalies. ResAD~\cite{resad2024} learns class-generalizable anomaly representations through residual features.

Large vision-language models also support anomaly detection and reasoning through normal reference images. IADGPT~\cite{iadgpt2025} uses in-context learning to unify industrial anomaly detection, localization, and reasoning. VD-DeepStack~\cite{vddeepstack2026} conditions language reasoning on dense query--reference visual differences injected at multiple decoder depths. 

Normality modeling has also been explored in video anomaly detection through appearance--motion prototypes in AMP-Net~\cite{liu2024ampnet} and scene-debiasing representation learning in CRCL~\cite{liu2025crcl}. Zhao et al.~\cite{zhao2026semivad} further combine video descriptions with visual features to model normality under limited anomaly supervision.

Auxiliary-data-based methods learn anomaly knowledge that can transfer to unseen targets. AnomalyCLIP~\cite{anomalyclip} learns object-agnostic normal and abnormal prompts through joint global and local supervision for zero-shot transfer, while AA-CLIP~\cite{aaclip2025} constructs anomaly-aware text anchors before aligning visual features to them. Anchor and Adapt combines auxiliary semantic learning and normal-only target adaptation in a two-stage framework. It transfers learned normal and abnormal anchors to the target product and keeps them fixed while learning an additional class-specific normal branch. The inherited and adapted normal branches provide complementary evidence, with anchor regularization guiding adaptation relative to both sets of references.

\section{Preliminaries}
\label{sec:preliminary}

\subsection{CLIP and Vision-Language Transfer}
\label{subsec:clip}

CLIP~\cite{clip} learns a shared embedding space for images and text through large-scale contrastive pretraining. Let $f(\cdot)$ and $g(\cdot)$ denote the image encoder and text encoder, respectively. Given an image $\mathbf{x}$ and a set of text prompts $\{\mathbf{t}_i\}_{i=1}^{N}$, CLIP predicts the probability that $\mathbf{x}$ belongs to the class represented by $\mathbf{t}_y$ as
\begin{equation}
p(y|\mathbf{x}) =
\frac{
\exp\left(\langle f(\mathbf{x}), g(\mathbf{t}_y)\rangle / \tau\right)
}{
\sum_{i=1}^{N}
\exp\left(\langle f(\mathbf{x}), g(\mathbf{t}_i)\rangle / \tau\right)
},
\label{eq:clip_prob}
\end{equation}
where $\langle \cdot,\cdot \rangle$ denotes cosine similarity and $\tau$ is the temperature. Since the classifier is constructed from text prompts rather than fixed category-specific parameters, CLIP can transfer to unseen categories by replacing the textual descriptions. This property makes CLIP attractive for few-shot and open-category recognition tasks.

\subsection{Prompt Learning}
\label{subsec:prompt_learning}

In standard zero-shot classification, CLIP usually relies on manually designed prompts, such as ``a photo of a \texttt{[class]}''. However, prompt wording can strongly affect performance, and searching for effective prompts is often time-consuming. Prompt learning addresses this issue by replacing hand-crafted context words with learnable vectors. For example, CoOp \cite{CoOp} represents a prompt as
\begin{equation}
\mathbf{P}
=
[\mathbf{p}_1][\mathbf{p}_2]\ldots[\mathbf{p}_{L}][\text{class}],
\label{eq:prompt_learning}
\end{equation}
where $\mathbf{p}_1,\ldots,\mathbf{p}_L$ are learnable context tokens and $[\text{class}]$ is the class name. By optimizing only a small number of prompt parameters while keeping the pretrained vision-language model frozen, prompt learning provides a parameter-efficient way to adapt CLIP to downstream tasks with limited data.

\subsection{CLIP for Few-Shot Industrial Anomaly Detection}
\label{subsec:clip_fsiad}

CLIP-based anomaly detection treats normality and abnormality as textual concepts. Instead of classifying an image into object categories, these methods compare visual features with normal and abnormal text features, and use their similarity difference to estimate anomaly scores. For image-level detection, the global image feature is compared with normal/abnormal text prompts. For pixel-level localization, patch features from intermediate visual layers are compared with the same text features to produce anomaly maps. Methods such as AnomalyCLIP~\cite{anomalyclip} and PromptAD~\cite{promptad} follow this paradigm by learning or designing prompts for normal and abnormal states, and some methods further enhance local visual features with V-V attention mechanisms inspired by CLIP-Surgery \cite{clip-surg}.

\section{Method}
\label{sec:method}

\subsection{Overview}
\label{subsec:overview}
Anchor and Adapt separates the acquisition of transferable anomaly semantics from adaptation to target normal appearance, as illustrated in Figure~\ref{fig:framework}. Stage I uses annotated auxiliary data to learn generic normal and abnormal anchors through object-agnostic prompt learning~\cite{anomalyclip}. Stage II retains these anchors and learns an additional class-specific normal branch from the few target normal images. The inherited and adapted normal branches jointly represent normality, while text-anchor regularization guides adaptation relative to the fixed references.

\subsection{Stage I: Auxiliary Anchor Pretraining}
\label{subsec:stage1}

Following the object-agnostic prompt formulation of AnomalyCLIP~\cite{anomalyclip}, we pretrain generic normal and abnormal anchors on annotated auxiliary data. The visual and text encoders $f(\cdot)$ and $g(\cdot)$ remain frozen, and their output features are $\ell_2$-normalized. We use $R_n$ normal and $R_a$ abnormal prompt contexts:
\begin{equation}
\begin{aligned}
\mathbf{P}_{n}^{r} &= [\mathbf{V}^{r}_{1}]\ldots[\mathbf{V}^{r}_{E}][\text{object}],
\quad r=1,\ldots,R_n, \\
\mathbf{P}_{a}^{s} &= [\mathbf{W}^{s}_{1}]\ldots[\mathbf{W}^{s}_{E}][\text{damaged}][\text{object}], \\
&\hspace{7em} s=1,\ldots,R_a,
\end{aligned}
\label{eq:stage1_prompts}
\end{equation}
where $\mathbf{V}$ and $\mathbf{W}$ are learnable tokens. Let $\mathbf{w}_b^r=g(\mathbf{P}_b^r)$ for $b\in\{n,a\}$. For a visual feature $\mathbf{u}$, multiple anchors are aggregated by normalized log-sum-exp:
\begin{equation}
\ell_b(\mathbf{u})=
\log\left[\frac{1}{R_b}\sum_{r=1}^{R_b}
\exp\left(\langle\mathbf{u},\mathbf{w}_b^r\rangle/\tau\right)\right].
\label{eq:stage1_logits}
\end{equation}
Here, $\tau$ is the temperature and $R_b$ denotes the normal or abnormal anchor count.

We follow AnomalyCLIP's global and local supervision scheme~\cite{anomalyclip}:
\begin{equation}
\mathcal{L}_{\mathrm{stage1}}
=\mathcal{L}_{\mathrm{img}}+\lambda\mathcal{L}_{\mathrm{pix}}.
\label{eq:stage1_loss}
\end{equation}
Here, $\lambda$ balances global and local supervision, and $\mathcal{L}_{\mathrm{img}}$ is cross-entropy between the image label and the binary softmax of $[\ell_n(\mathbf{m}),\ell_a(\mathbf{m})]$ for the global feature $\mathbf{m}$. The pixel-level term is
\begin{equation}
\begin{aligned}
\mathcal{L}_{\mathrm{pix}}
=\sum_{\ell\in\mathcal{L}_{\mathrm{feat}}}\Big[& F(\mathbf{S}^{\ell},\mathbf{G})
+D(\mathbf{S}^{\ell}_{a},\mathbf{G}) \\
&+D(\mathbf{S}^{\ell}_{n},\mathbf{1}-\mathbf{G})\Big],
\end{aligned}
\label{eq}
\end{equation}
where $\mathcal{L}_{\mathrm{feat}}$ denotes the selected visual layers and $\mathbf{S}^{\ell}$ is the two-channel normal/abnormal probability map computed from the patch-level logits at layer $\ell$ and aligned with the anomaly mask $\mathbf{G}$. The functions $F$ and $D$ denote focal~\cite{focal} and Dice~\cite{dict} losses. The pretrained prompts supply the fixed references, and the normal contexts initialize the class-specific normal branch in Stage II.

\subsection{Stage II: Asymmetric Target Prompt Adaptation}
\label{subsec:stage2_adaptation}

Following the normal-only target learning setting of PromptAD~\cite{promptad}, adaptation uses only the few normal images of each unseen product. We retain both auxiliary-learned anchor sets as fixed references and learn an additional class-specific normal branch. Target normality is represented jointly by this adapted branch and the inherited generic normal branch.

\subsubsection{Class-specific normal prompt construction}
For each target class, we maintain two types of normal prompts.
The first type is the inherited generic normal prompt, which directly uses the transferable normal context learned in Stage I:
\begin{equation}
\mathbf{P}_{n}^{g,r}
=
[\mathbf{V}^{r}_{1}]\ldots[\mathbf{V}^{r}_{E}][\text{object}],
\quad r=1,\ldots,R_n .
\label{eq:stage2_generic_prompt}
\end{equation}
The second type is the class-specific normal prompt, which adapts the transferable normal context to the target product:
\begin{equation}
\begin{aligned}
\mathbf{P}_{n}^{c,q}
={}&[\mathbf{V}^{q}_{1}+\Delta\mathbf{V}^{q}_{1}]\ldots
[\mathbf{V}^{q}_{E}+\Delta\mathbf{V}^{q}_{E}] \\
&[\mathbf{N}^{q}_{1}]\ldots[\mathbf{N}^{q}_{J}][\text{class}],
\quad q=1,\ldots,R_c .
\end{aligned}
\label{eq:stage2_class_prompt}
\end{equation}
Here, $\mathbf{V}^{q}$ is copied from the transferable normal context, $\Delta\mathbf{V}^{q}$ is a low-rank residual adaptation, $\mathbf{N}^{q}$ denotes newly learnable tokens, and $[\text{class}]$ is replaced by the target class name.
The low-rank residual is parameterized as
\begin{equation}
\begin{aligned}
\Delta\mathbf{V}^{q}&=\mathbf{A}^{q}\mathbf{B}^{q},\\
\mathbf{A}^{q}&\in\mathbb{R}^{E\times\rho},\quad
\mathbf{B}^{q}\in\mathbb{R}^{\rho\times d},\quad\rho\ll d ,
\end{aligned}
\label{eq:stage2_low_rank}
\end{equation}
where $d$ is the prompt embedding dimension and $\rho$ is the factorization width, which controls the capacity of the residual adaptation.

The abnormal prompts are inherited from Stage I and kept frozen:
\begin{equation}
\begin{aligned}
\mathbf{P}_{a}^{s}
&=[\mathbf{W}^{s}_{1}]\ldots[\mathbf{W}^{s}_{E}][\text{damaged}][\text{object}],\\
&\hspace{7em}s=1,\ldots,R_a .
\end{aligned}
\label{eq:stage2_abnormal_prompt}
\end{equation}
During the adaptation stage, the VLM backbone, the generic normal prompts, and the abnormal prompts are frozen.
Only the class-specific adaptation parameters, including the low-rank residuals and the newly learnable tokens, are optimized.

The corresponding text features are denoted as
\begin{equation}
\mathbf{w}_{n}^{g,r}=g(\mathbf{P}_{n}^{g,r}),
\quad
\mathbf{w}_{n}^{c,q}=g(\mathbf{P}_{n}^{c,q}),
\quad
\mathbf{w}_{a}^{s}=g(\mathbf{P}_{a}^{s}).
\end{equation}

\subsubsection{Dual-normal evidence}
We define the prompt-based logits for a visual feature $\mathbf{u}$. For image-level adaptation, $\mathbf{u}$ denotes the global image feature $\mathbf{m}$; for pixel-level adaptation, it denotes a patch feature $\mathbf{p}_{l,j}$.
The normal evidence is aggregated from both the inherited generic normal prompts and the adapted class-specific normal prompts:
\begin{equation}
\begin{aligned}
\ell_{n}^{*}(\mathbf{u})=\log\frac{1}{R_n+R_c}
\Bigg[&\sum_{r=1}^{R_n}
\exp\left(\langle\mathbf{u},\mathbf{w}_{n}^{g,r}\rangle/\tau\right)\\
+&\sum_{q=1}^{R_c}
\exp\left(\langle\mathbf{u},\mathbf{w}_{n}^{c,q}\rangle/\tau\right)\Bigg],
\end{aligned}
\label{eq:stage2_dual_normal_logit}
\end{equation}
while the abnormal evidence is computed from the frozen abnormal prompts:
\begin{equation}
\ell_{a}^{*}(\mathbf{u})
=
\log
\frac{1}{R_a}
\sum_{s=1}^{R_a}
\exp\left(
\langle \mathbf{u}, \mathbf{w}_{a}^{s}\rangle/\tau
\right).
\label{eq:stage2_abnormal_logit}
\end{equation}
The dual-normal formulation allows the inherited generic normal prompts to preserve transferable normality, while the class-specific normal prompts adapt to the appearance of the target product.

\subsubsection{Text-anchor regularization}
We constrain the class-specific normal prompts relative to both sets of inherited anchors. For each adapted prompt, text-anchor regularization encourages consistency with the generic normal reference and separation from the fixed abnormal reference:
\begin{equation}
\begin{aligned}
\mathcal{L}_{\mathrm{anchor}}
=\frac{1}{R_c}\sum_{q=1}^{R_c}\Big[&
\max\left(0,m_n-\max_r
\langle\mathbf{w}_{n}^{c,q},\mathbf{w}_{n}^{g,r}\rangle\right)\\
+&\max\left(0,\max_s
\langle\mathbf{w}_{n}^{c,q},\mathbf{w}_{a}^{s}\rangle-m_a\right)\Big],
\end{aligned}
\label{eq:text_anchor_reg}
\end{equation}
where $m_n$ and $m_a$ denote the normal-anchor and abnormal-anchor margins, respectively.
The first term encourages consistency with the transferable normal prior, while the second encourages separation from the abnormal anchors.

\subsubsection{Image-level prompt adaptation}
For image-level anomaly detection, we apply the prompt-based logits to the global image feature $\mathbf{m}=f(x)$.
Since all target training images are normal, the image-level target adaptation loss is
\begin{equation}
\mathcal{L}_{\mathrm{img}}
=
-\mathbb{E}_{\mathbf{m}}
\log
\frac{
\exp(\ell_{n}^{*}(\mathbf{m}))
}{
\exp(\ell_{n}^{*}(\mathbf{m}))
+
\exp(\ell_{a}^{*}(\mathbf{m}))
}.
\label{eq:stage2_image_ce}
\end{equation}
The image-level objective is
\begin{equation}
\mathcal{L}_{\mathrm{stage2}}^{\mathrm{img}}
=
\mathcal{L}_{\mathrm{img}}
+
\lambda_{\mathrm{anchor}}
\mathcal{L}_{\mathrm{anchor}},
\label{eq:stage2_img_loss}
\end{equation}
where $\lambda_{\mathrm{anchor}}$ controls the strength of the text-anchor regularization.

\subsubsection{Pixel-level prompt adaptation}
For pixel-level anomaly localization, the target-stage supervision is applied to local patch features rather than global image features.
Let $\mathbf{p}_{l,j}$ denote the $j$-th patch feature extracted from the $l$-th selected visual layer.
Since all target training images are normal, these patch features are treated as local normal evidence.
We therefore use the following patch-level loss based on the dual-normal logits:
\begin{equation}
\mathcal{L}_{\mathrm{patch}}
=
-\mathbb{E}_{l,j}
\log
\frac{
\exp(\ell_{n}^{*}(\mathbf{p}_{l,j}))
}{
\exp(\ell_{n}^{*}(\mathbf{p}_{l,j}))
+
\exp(\ell_{a}^{*}(\mathbf{p}_{l,j}))
}.
\label{eq:stage2_patch_ce}
\end{equation}
The pixel-level objective is
\begin{equation}
\mathcal{L}_{\mathrm{stage2}}^{\mathrm{pix}}
=
\mathcal{L}_{\mathrm{patch}}
+
\lambda_{\mathrm{anchor}}
\mathcal{L}_{\mathrm{anchor}}.
\label{eq:stage2_pix_loss}
\end{equation}
Thus, image-level and pixel-level adaptation share the same prompt formulation and text-anchor regularization, but differ in the visual features used for supervision: global image features for image-level anomaly detection and patch features for pixel-level anomaly localization.

\subsection{Inference}
\label{subsec:inference}

At inference time, the dual-normal representation and the fixed abnormal anchors provide prompt-guided anomaly scores.

\subsubsection{Pixel-level anomaly scoring}
For anomaly localization, we extract patch features
$\{\mathbf{p}_{\ell,i}\}_{i=1}^{N_{\ell}}$ from selected visual layers
$\ell\in\mathcal{L}_{\mathrm{feat}}$.
For each patch feature, we first compute the prompt-guided logit difference between the abnormal and normal evidence:
\begin{equation}
d_{\mathrm{text}}^{\ell}(i)
=
\ell_{a}^{*}(\mathbf{p}_{\ell,i})
-
\ell_{n}^{*}(\mathbf{p}_{\ell,i}),
\label{eq:text_patch_diff}
\end{equation}
where $\ell_{n}^{*}(\cdot)$ and $\ell_{a}^{*}(\cdot)$ are the dual-normal and abnormal logits defined in Stage II.
The scores from different layers are aggregated by a normalized log-sum-exp operator:
\begin{equation}
d_{\mathrm{text}}(i)
=
\tau
\log
\frac{1}{|\mathcal{L}_{\mathrm{feat}}|}
\sum_{\ell\in\mathcal{L}_{\mathrm{feat}}}
\exp
\left(
d_{\mathrm{text}}^{\ell}(i)/\tau
\right),
\label{eq:text_layer_fusion}
\end{equation}
where $\tau$ is the layer-fusion temperature.
The text-guided anomaly map is then obtained by applying a sigmoid function, reshaping the patch scores to the spatial grid, and upsampling them to the evaluation resolution:
\begin{equation}
A_{\mathrm{text}}
=
\operatorname{Up}
\left(
\operatorname{Reshape}
\left(
\sigma(d_{\mathrm{text}})
\right)
\right).
\label{eq:text_map}
\end{equation}

Following the normal-reference matching in PromptAD~\cite{promptad}, we construct a visual normal memory $\mathcal{B}_{\ell}$ from the $\ell_2$-normalized patch features of the few-shot normal images.
For each test patch, the visual-reference anomaly score is computed as its nearest-neighbor normalized cosine distance to the normal memory:
\begin{equation}
d_{\mathrm{vis}}^{\ell}(i)
=
\min_{\mathbf{b}\in\mathcal{B}_{\ell}}
\frac{
1-
\left\langle
\mathbf{p}_{\ell,i},\mathbf{b}
\right\rangle
}{2},
\label{eq:visual_patch_score}
\end{equation}
where both $\mathbf{p}_{\ell,i}$ and $\mathbf{b}$ are $\ell_2$-normalized patch features.
A larger $d_{\mathrm{vis}}^{\ell}(i)$ indicates that the test patch is less similar to its nearest normal reference patch and is therefore more likely to be anomalous.
The normalized visual-reference scores are averaged over the selected feature layers and upsampled to the evaluation resolution:
\begin{equation}
A_{\mathrm{vis}}
=
\operatorname{Up}
\left(
\operatorname{Reshape}
\left(
\frac{1}{|\mathcal{L}_{\mathrm{feat}}|}
\sum_{\ell\in\mathcal{L}_{\mathrm{feat}}}
d_{\mathrm{vis}}^{\ell}
\right)
\right),
\label{eq:visual_map}
\end{equation}
where the same feature-layer set $\mathcal{L}_{\mathrm{feat}}$ is used for the visual memory branch.

Before fusion, the text-guided and visual-reference maps are normalized:
\begin{equation}
\hat{A}
=
\frac{
A-\min(A)
}{
\max(A)-\min(A)+\epsilon
},
\label{eq:map_norm}
\end{equation}
where $\epsilon$ is a small constant for numerical stability.
The final pixel-level anomaly map is computed by weighted averaging:
\begin{equation}
A_{\mathrm{pix}}
=
\frac{
\alpha_{\mathrm{text}}\hat{A}_{\mathrm{text}}
+
\alpha_{\mathrm{vis}}\hat{A}_{\mathrm{vis}}
}{
\alpha_{\mathrm{text}}+\alpha_{\mathrm{vis}}
}.
\label{eq:pixel_weighted_fusion}
\end{equation}
In our implementation, we use equal weights for the two branches, i.e.,
$\alpha_{\mathrm{text}}=\alpha_{\mathrm{vis}}=1$.

\subsubsection{Image-level anomaly scoring}
For image-level anomaly detection, we aggregate prompt-guided semantic evidence and visual-reference evidence. Given the global image feature $\mathbf{m}$, the text-based anomaly score is computed as
\begin{equation}
S_{\mathrm{text}}(x)
=
\frac{
\exp(\ell_a^{*}(\mathbf{m}))
}{
\exp(\ell_n^{*}(\mathbf{m}))+
\exp(\ell_a^{*}(\mathbf{m}))
}.
\label{eq:image_text_score}
\end{equation}
This score measures the image-level affinity to the frozen abnormal anchor relative to the dual-normal evidence.

Using the visual-reference map $A_{\mathrm{vis}}$ in Eq.~\ref{eq:visual_map}, we summarize image-level visual evidence by averaging its top-$K$ spatial responses:
\begin{equation}
S_{\mathrm{vis}}(x)
=
\frac{1}{K}
\sum_{r \in \operatorname{TopK}(A_{\mathrm{vis}}, K)} r.
\label{eq:image_visual_score}
\end{equation}

When the fused pixel-level anomaly map is used for image-level scoring, we summarize it by averaging its top-$K$ responses:
\begin{equation}
S_{\mathrm{pix}}(x)
=
\frac{1}{K}
\sum_{r \in \operatorname{TopK}(A_{\mathrm{pix}},K)}
r .
\label{eq:image_pixel_topk_score}
\end{equation}

We use the harmonic-fusion form of PromptAD~\cite{promptad} to combine the image-level scores:
\begin{equation}
\begin{aligned}
S_{\mathrm{img}}(x)=\Bigg(&\frac{1}{S_{\mathrm{text}}(x)+\epsilon}
+\frac{1}{S_{\mathrm{vis}}(x)+\epsilon}\\
&+\frac{1}{S_{\mathrm{pix}}(x)+\epsilon}\Bigg)^{-1}.
\end{aligned}
\label{eq:image_final_score}
\end{equation}
If the pixel-level score is not used, the last term is omitted.

\section{Experiments}
\label{sec:experiments}

\begin{table*}[t]
\caption{Few-shot anomaly detection and localization on VisA (\%). Bold and underlined values indicate the best and second-best results, respectively. A dash denotes an unreported result.}
\label{tab:main_visa}
\centering\footnotesize
\setlength{\tabcolsep}{2.3pt}\renewcommand{\arraystretch}{1.13}
\begin{tabular*}{\textwidth}{@{\extracolsep{\fill}}l*{12}{c}@{}}
\toprule
\multirow{2}{*}{Method} & \multicolumn{4}{c}{1-shot} & \multicolumn{4}{c}{2-shot} & \multicolumn{4}{c}{4-shot} \\
\cmidrule(lr){2-5}\cmidrule(lr){6-9}\cmidrule(lr){10-13}
& I-AUC & I-AUPR & P-AUC & PRO & I-AUC & I-AUPR & P-AUC & PRO & I-AUC & I-AUPR & P-AUC & PRO \\
\midrule
SPADE \cite{spade} & 79.5 & 82.0 & 95.6 & 84.1 & 80.7 & 82.3 & 96.2 & 85.7 & 81.7 & 83.4 & 96.6 & 87.3 \\
PaDiM \cite{padim} & 62.8 & 68.3 & 89.9 & \textemdash & 67.4 & 71.6 & 92.0 & \textemdash & 72.8 & 75.6 & 93.2 & \textemdash \\
PatchCore \cite{patchcore} & 79.9 & 82.8 & 95.4 & 80.5 & 81.6 & 84.8 & 96.1 & 82.6 & 85.3 & 87.5 & 96.8 & 84.9 \\
WinCLIP+ \cite{winclip} & 83.8 & 85.1 & 96.4 & 85.1 & 84.6 & 85.8 & 96.8 & 86.2 & 87.3 & 88.8 & 97.2 & 87.6 \\
RWDA \cite{rwda} & 83.4 & 85.8 & \textemdash & \textemdash & 85.6 & 87.5 & \textemdash & \textemdash & 86.6 & 88.4 & \textemdash & \textemdash \\
AnomalyGPT \cite{anomalygpt} & 87.4 & \textemdash & 96.2 & \textemdash & \underline{88.6} & \textemdash & 96.4 & \textemdash & 90.6 & \textemdash & 96.7 & \textemdash \\
PromptAD \cite{promptad} & 86.9 & \underline{88.4} & 96.7 & 85.1 & 88.3 & \underline{90.0} & 97.1 & 85.8 & 89.1 & \underline{90.8} & 97.4 & 86.2 \\
ResAD~\cite{resad2024} & \textemdash & \textemdash & \textemdash & \textemdash & 84.5 & \textemdash & 95.1 & \textemdash & \underline{90.8} & \textemdash & 97.5 & \textemdash \\
InCTRL~\cite{inctrl} & \textemdash & \textemdash & \textemdash & \textemdash & 85.8 & 87.7 & \textemdash & \textemdash & 87.7 & 90.2 & \textemdash & \textemdash \\
IIPAD~\cite{oneforall} & 85.4 & 87.5 & 96.9 & 87.3 & 86.7 & 88.6 & 97.2 & 87.9 & 88.3 & 89.6 & 97.4 & 88.3 \\

FiLo++ \cite{filo++} & \underline{88.3} & \textemdash & \textbf{97.3} & \textemdash & \underline{88.6} & \textemdash & \underline{97.5} & \textemdash & 89.8 & \textemdash & \textbf{97.9} & \textemdash \\
FastRef + WinCLIP~\cite{fastref} & 83.9 & \textemdash & 95.8 & \textemdash & 84.1 & \textemdash & 96.4 & \textemdash & 85.0 & \textemdash & 96.6 & \textemdash \\
FastRef + AnomalyDINO~\cite{fastref} & 84.6 & \textemdash & \underline{97.2} & \textemdash & 84.8 & \textemdash & 97.3 & \textemdash & 87.1 & \textemdash & \textbf{97.9} & \textemdash \\
FineGrainedAD~\cite{finegrainedad2026} & 84.2 & \textemdash & 96.4 & \textemdash & 86.2 & \textemdash & 97.0 & \textemdash & 87.0 & \textemdash & 97.5 & \textemdash \\
AnoPLe~\cite{anople2026} & 86.0 & 87.7 & 97.0 & \underline{87.5} & 86.9 & 88.6 & 97.3 & \underline{88.1} & 87.5 & 89.2 & 97.5 & \underline{88.5} \\

\midrule
\textbf{Ours} & \textbf{90.3} & \textbf{91.6} & \textbf{97.3} & \textbf{89.7} & \textbf{91.4} & \textbf{92.5} & \textbf{97.6} & \textbf{90.2} & \textbf{92.1} & \textbf{93.0} & \underline{97.8} & \textbf{91.1} \\
\bottomrule
\end{tabular*}
\end{table*}

\begin{table*}[t]
\caption{Few-shot anomaly detection and localization on MVTec-AD (\%). Bold and underlined values indicate the best and second-best results, respectively. A dash denotes an unreported result.}
\label{tab:main_mvtec}
\centering\footnotesize
\setlength{\tabcolsep}{2.3pt}\renewcommand{\arraystretch}{1.13}
\begin{tabular*}{\textwidth}{@{\extracolsep{\fill}}l*{12}{c}@{}}
\toprule
\multirow{2}{*}{Method} & \multicolumn{4}{c}{1-shot} & \multicolumn{4}{c}{2-shot} & \multicolumn{4}{c}{4-shot} \\
\cmidrule(lr){2-5}\cmidrule(lr){6-9}\cmidrule(lr){10-13}
& I-AUC & I-AUPR & P-AUC & PRO & I-AUC & I-AUPR & P-AUC & PRO & I-AUC & I-AUPR & P-AUC & PRO \\
\midrule
SPADE \cite{spade} & 81.0 & 90.6 & 91.2 & 83.9 & 82.9 & 91.7 & 92.0 & 85.7 & 84.8 & 92.5 & 92.7 & 87.0 \\
PaDiM \cite{padim} & 76.6 & 88.1 & 89.3 & \textemdash & 78.9 & 89.3 & 91.3 & \textemdash & 80.4 & 90.5 & 92.6 & \textemdash \\
PatchCore \cite{patchcore} & 83.4 & 92.2 & 92.0 & 79.7 & 86.3 & 93.8 & 93.3 & 82.3 & 88.8 & 94.5 & 94.3 & 84.3 \\
WinCLIP+ \cite{winclip} & 93.1 & 96.5 & 95.2 & 87.1 & 94.4 & 97.0 & 96.0 & 88.4 & 95.2 & 97.3 & 96.2 & 89.0 \\
RWDA \cite{rwda} & 93.3 & 96.7 & \textemdash & \textemdash & 94.0 & 96.9 & \textemdash & \textemdash & 94.5 & 97.1 & \textemdash & \textemdash \\
FastRecon \cite{fastrecon} & \textemdash & \textemdash & \textemdash & \textemdash & 91.0 & \textemdash & 95.9 & \textemdash & 94.2 & \textemdash & \underline{97.0} & \textemdash \\
AnomalyGPT \cite{anomalygpt} & 94.1 & \textemdash & 95.3 & \textemdash & 95.5 & \textemdash & 95.6 & \textemdash & 96.3 & \textemdash & 96.2 & \textemdash \\
PromptAD \cite{promptad} & 94.6 & 97.1 & 95.9 & 87.9 & 95.7 & 97.9 & 96.2 & 88.5 & 96.6 & \underline{98.5} & 96.5 & 90.5 \\
ResAD~\cite{resad2024} & \textemdash & \textemdash & \textemdash & \textemdash & 94.4 & \textemdash & 95.6 & \textemdash & 94.2 & \textemdash & 96.9 & \textemdash \\
InCTRL~\cite{inctrl} & \textemdash & \textemdash & \textemdash & \textemdash & 94.0 & 96.9 & \textemdash & \textemdash & 94.5 & 97.2 & \textemdash & \textemdash \\
IIPAD~\cite{oneforall} & 94.2 & 97.2 & \underline{96.4} & 89.8 & 95.7 & 97.9 & \textbf{96.7} & 90.3 & 96.1 & 98.1 & \underline{97.0} & 91.2 \\
FiLo++ \cite{filo++} & 95.0 & \textemdash & 96.2 & \textemdash & 95.8 & \textemdash & 96.5 & \textemdash & 96.3 & \textemdash & 96.6 & \textemdash \\
FastRef + WinCLIP~\cite{fastref} & 93.8 & \textemdash & 95.7 & \textemdash & 93.9 & \textemdash & 96.2 & \textemdash & 95.5 & \textemdash & 96.7 & \textemdash \\
FastRef + AnomalyDINO~\cite{fastref} & \underline{97.0} & \textemdash & \textbf{96.5} & \textemdash & \underline{97.2} & \textemdash & \textbf{96.7} & \textemdash & \underline{97.4} & \textemdash & 96.8 & \textemdash \\
FineGrainedAD~\cite{finegrainedad2026} & 91.0 & \textemdash & 95.5 & \textemdash & 93.7 & \textemdash & 96.1 & \textemdash & 94.5 & \textemdash & 96.7 & \textemdash \\
AnoPLe~\cite{anople2026} & 94.5 & \underline{97.4} & 96.0 & \textbf{90.8} & 95.8 & \underline{98.0} & 96.3 & \textbf{91.4} & 96.4 & 98.4 & 96.5 & \underline{91.3} \\

\midrule
\textbf{Ours} & \textbf{97.6} & \textbf{98.8} & \underline{96.4} & \underline{90.4} & \textbf{97.9} & \textbf{99.1} & \underline{96.6} & \underline{91.0} & \textbf{98.3} & \textbf{99.4} & \textbf{97.3} & \textbf{92.3} \\
\bottomrule
\end{tabular*}
\end{table*}

\begin{table*}[t]
\caption{Few-shot anomaly detection and localization on MPDD (\%). Bold and underlined values indicate the best and second-best results, respectively. A dash denotes an unreported result.}
\label{tab:main_mpdd}
\centering\footnotesize
\setlength{\tabcolsep}{2.3pt}\renewcommand{\arraystretch}{1.13}
\begin{tabular*}{\textwidth}{@{\extracolsep{\fill}}l*{12}{c}@{}}
\toprule
\multirow{2}{*}{Method} & \multicolumn{4}{c}{1-shot} & \multicolumn{4}{c}{2-shot} & \multicolumn{4}{c}{4-shot} \\
\cmidrule(lr){2-5}\cmidrule(lr){6-9}\cmidrule(lr){10-13}
& I-AUC & I-AUPR & P-AUC & PRO & I-AUC & I-AUPR & P-AUC & PRO & I-AUC & I-AUPR & P-AUC & PRO \\
\midrule
PatchCore~\cite{patchcore} & 59.2 & \textemdash & 78.5 & \textemdash & 59.6 & \textemdash & 79.2 & \textemdash & 79.9 & \textemdash & 79.8 & \textemdash \\
WinCLIP+~\cite{winclip} & 68.2 & \textemdash & 92.6 & \textemdash & 69.3 & \textemdash & 94.7 & \textemdash & 75.2 & \textemdash & 96.0 & \textemdash \\
FastRecon~\cite{fastrecon} & \textemdash & \textemdash & \textemdash & \textemdash & 73.7 & \textemdash & 97.0 & \textemdash & 79.9 & \textemdash & 97.6 & \textemdash \\
AnomalyGPT~\cite{anomalygpt,dictas} & 72.4 & \underline{75.9} & 96.6 & \underline{89.9} & 79.7 & \underline{81.1} & 97.3 & \underline{91.8} & \underline{85.9} & \textbf{89.0} & \underline{97.7} & \underline{93.2} \\
PromptAD~\cite{promptad} & \underline{80.7} & \textemdash & 96.2 & \textemdash & \underline{85.3} & \textemdash & 97.2 & \textemdash & \textbf{87.2} & \textemdash & 97.3 & \textemdash \\

FastRef + WinCLIP~\cite{fastref} & 72.5 & \textemdash & 96.9 & \textemdash & 76.0 & \textemdash & 97.3 & \textemdash & 82.0 & \textemdash & 97.6 & \textemdash \\
FastRef + AnomalyDINO~\cite{fastref} & 76.7 & \textemdash & \underline{97.1} & \textemdash & 78.4 & \textemdash & \underline{97.4} & \textemdash & 82.1 & \textemdash & 97.6 & \textemdash \\
\midrule
\textbf{Ours} & \textbf{82.0} & \textbf{82.7} & \textbf{97.5} & \textbf{92.8} & \textbf{85.8} & \textbf{86.3} & \textbf{97.7} & \textbf{93.4} & \textbf{87.2} & \underline{88.0} & \textbf{98.0} & \textbf{94.2} \\
\bottomrule
\end{tabular*}

\end{table*}

We conduct our primary few-shot detection and localization experiments on VisA and MVTec-AD under 1-shot, 2-shot, and 4-shot settings. We additionally evaluate on MPDD to examine the transferability of the proposed framework to metal-part inspection. Ablation studies and qualitative visualizations examine the proposed adaptation design.

\subsection{Experimental Setup}
\subsubsection{Datasets and Evaluation Metrics}
VisA~\cite{visa} and MVTec-AD~\cite{Mvtec}, which contain 12 and 15 product categories, respectively, are commonly used to evaluate few-shot industrial anomaly detection~\cite{promptad,oneforall,finegrainedad2026,anople2026}. Following this practice, we use them as our primary benchmarks. We additionally evaluate on MPDD~\cite{mpdd}, which contains six categories of metal parts, to assess the transferability of our method beyond these two commonly used benchmarks.
Following the few-shot protocols of PromptAD~\cite{promptad} and IIPAD~\cite{oneforall}, for each target category in VisA and MVTec-AD, we sample $k\in\{1,2,4\}$ normal images from the training split to form the support set. The same support images are used for Stage-II normal-prompt adaptation and visual-memory construction. The full test split serves as the query set and remains fixed across all shot settings. The support and query sets are disjoint, and query images and labels are not used to optimize target prompts.

We report image- and pixel-level AUROC, image-level AUPR, and region-level PRO. The table headings I-AUC and P-AUC denote image- and pixel-level AUROC, respectively. On MPDD, the I-AUPR and PRO columns report average precision (AP) and area under the per-region overlap curve (AUPRO), respectively. All results are expressed as percentages; higher values indicate better performance.

\subsubsection{Implementation Details}

We use CLIP ViT-L/14@336px as the backbone.
Multi-level visual features are extracted from layers 6, 12, 18, and 24.
In Stage I, we set the context length to $E=12$ and learn $8$ normal prompts and $8$ abnormal prompts from the auxiliary anomaly dataset.
For cross-dataset evaluation, Stage I uses VisA as the auxiliary dataset for MVTec-AD, and MVTec-AD for VisA and MPDD. Stage II adapts the normal prompts using only the few normal training images from each target category; target test data are excluded from both stages.
In Stage II, we introduce $8$ adapted class-specific normal prompts, with $J=2$ newly learnable tokens and a low-rank residual factorization width of $\rho=8$.  The top-$K$ parameter is set to $K=7$.
The weight of the text-anchor regularization is set to $\lambda_{\mathrm{anchor}}=0.1$.
The first stage is trained for 15 epochs, and the second stage is trained for 150 epochs.
All experiments are run on a single NVIDIA Tesla A800 GPU.

\subsection{Main Results}
\label{subsec:sota_comparison}

\subsubsection{Comparison methods}
Tables~\ref{tab:main_visa}--\ref{tab:main_mpdd} compare our method with feature-based and VLMs baselines using their reported results. FineGrainedAD~\cite{finegrainedad2026} uses category-specific predefined anomaly descriptions, while AnoPLe~\cite{anople2026} uses synthetic anomalous images in joint multi-class training. Our framework learns anomaly references from real auxiliary data and adapts normal prompts for each target category, requiring neither category-specific anomaly templates nor synthetic anomaly generation.

\subsubsection{Image-level anomaly detection}
Our method achieves the highest image-level AUROC and AUPR across all three shot settings in Tables~\ref{tab:main_visa} and~\ref{tab:main_mvtec}. The improvements are more pronounced on VisA, which presents a more challenging image-level detection benchmark. Our VisA AUROC scores of 90.3, 91.4, and 92.1 exceed the next-best results by 2.0, 2.8, and 1.3 percentage points, respectively. On MVTec-AD, leading methods already achieve high AUROC scores of 97.0--97.4, leaving less room for further improvement. Our method further improves these results by 0.6, 0.7, and 0.9 points, reaching 97.6, 97.9, and 98.3 AUROC, respectively. Image-level AUPR reaches 93.0 on VisA and 99.4 on MVTec-AD at 4 shots.

\subsubsection{Pixel-level anomaly localization}
On VisA, our pixel-level AUROC of 97.3 ties the best result at 1 shot, 97.6 is the best at 2 shots, and 97.8 is second-best at 4 shots. Our VisA PRO scores of 89.7, 90.2, and 91.1 are the highest across all three settings in Table~\ref{tab:main_visa}. On MVTec-AD, our method achieves the second-best pixel-level AUROC and PRO at 1 and 2 shots, and the best values at 4 shots. The 4-shot scores of 97.3 AUROC and 92.3 PRO exceed the next-best results by 0.3 and 1.0 percentage points, respectively.

\subsubsection{Additional evaluation on MPDD}
Table~\ref{tab:main_mpdd} evaluates transfer from MVTec-AD to MPDD. Our image-level AUROC scores are 82.0, 85.8, and 87.2 at 1, 2, and 4 shots, exceeding PromptAD by 1.3 and 0.5 points in the first two settings and matching it at 4 shots. Image-level AUPR is highest at 1 and 2 shots and second-best at 4 shots. Our method also achieves the highest pixel-level AUROC and PRO across all three settings, reaching 98.0 and 94.2 at 4 shots. These results extend the transfer evaluation to metal-part inspection.

\subsection{Ablation Study}
\label{subsec:ablation}

\begin{table*}[t]
\centering
\begin{minipage}{0.82\textwidth}
\centering
\caption{Controlled comparisons of anchor construction and target adaptation on VisA (\%). The Stage-I-only baseline is evaluated zero-shot; all other variants use two normal target images per category. Bold indicates the best result.}
\label{tab:ablation_prompt_design}
\footnotesize
\setlength{\tabcolsep}{5pt}
\renewcommand{\arraystretch}{1.18}
\begin{tabular*}{\linewidth}{@{\extracolsep{\fill}}lcccc@{}}
\toprule
\multirow{2}{*}{Variant} & \multicolumn{2}{c}{Image-level} & \multicolumn{2}{c}{Pixel-level} \\
\cmidrule(lr){2-3}\cmidrule(lr){4-5}
& I-AUC & I-AUPR & P-AUC & PRO \\
\midrule
Stage-I only (zero-shot) & 83.9 & 86.3 & 95.3 & 88.4 \\
Stage-I anchors + normal memory & 87.6 & 89.1 & 97.1 & 89.9 \\
\midrule
Manual prompts + target adaptation & 87.3 & 89.1 & 97.2 & 90.0 \\
Constrained abnormal-anchor adaptation & 91.1 & 92.0 & 97.4 & 89.8 \\
w/o generic normal branch & 90.6 & 92.0 & 97.2 & 89.8 \\
w/o low-rank residual & 90.7 & 92.0 & 97.3 & 90.0 \\
w/o anchor regularization & 90.3 & 91.7 & 97.3 & 89.7 \\
\midrule
\textbf{Full model} & \textbf{91.4} & \textbf{92.5} & \textbf{97.6} & \textbf{90.2} \\
\bottomrule
\end{tabular*}
\end{minipage}

\par\vspace{8pt}

\begin{minipage}[t]{0.485\textwidth}
\centering
\caption{
Sensitivity analysis of the number of newly learnable tokens $J$ on VisA under the 2-shot setting (\%). Bold indicates the best result.
}
\label{tab:sensitivity_tokens}
\footnotesize\setlength{\tabcolsep}{2.5pt}
\renewcommand{\arraystretch}{1.12}
\begin{tabular*}{\columnwidth}{@{\extracolsep{\fill}}ccccc@{}}
\toprule
Number of tokens $J$ & I-AUROC & I-AUPR &P-AUROC & PRO \\
\midrule
$0$ & 89.4 & 90.4 & 97.5 & 90.1 \\
$2$ & \textbf{91.4} & \textbf{92.5} & \textbf{97.6} & \textbf{90.2} \\
$4$ & 90.9 & 92.3 & 97.5 & \textbf{90.2} \\
$8$ & 91.0 & 92.4 & 97.5 & \textbf{90.2} \\
\bottomrule
\end{tabular*}
\end{minipage}
\hfill
\begin{minipage}[t]{0.485\textwidth}
\centering
\caption{
Sensitivity analysis of the low-rank residual factorization width $\rho$ on VisA under the 2-shot setting (\%). Bold indicates the best result.
}
\label{tab:sensitivity_rank}
\footnotesize\setlength{\tabcolsep}{2.5pt}
\renewcommand{\arraystretch}{1.12}
\begin{tabular*}{\columnwidth}{@{\extracolsep{\fill}}ccccc@{}}
\toprule
Residual factorization width $\rho$ & I-AUROC & I-AUPR & P-AUROC & PRO \\
\midrule
$2$ & 90.9 & 92.3 & 97.4 & 90.1 \\
$4$ & 91.1 & 92.4 & 97.5 & \textbf{90.2} \\
$8$ & \textbf{91.4} & \textbf{92.5} & \textbf{97.6} & \textbf{90.2} \\
$12$ & 90.7 & 92.1 & 97.3 & 90.1 \\
\bottomrule
\end{tabular*}
\end{minipage}
\end{table*}

\begin{figure*}[t]
  \centering
  \includegraphics[width=\textwidth]{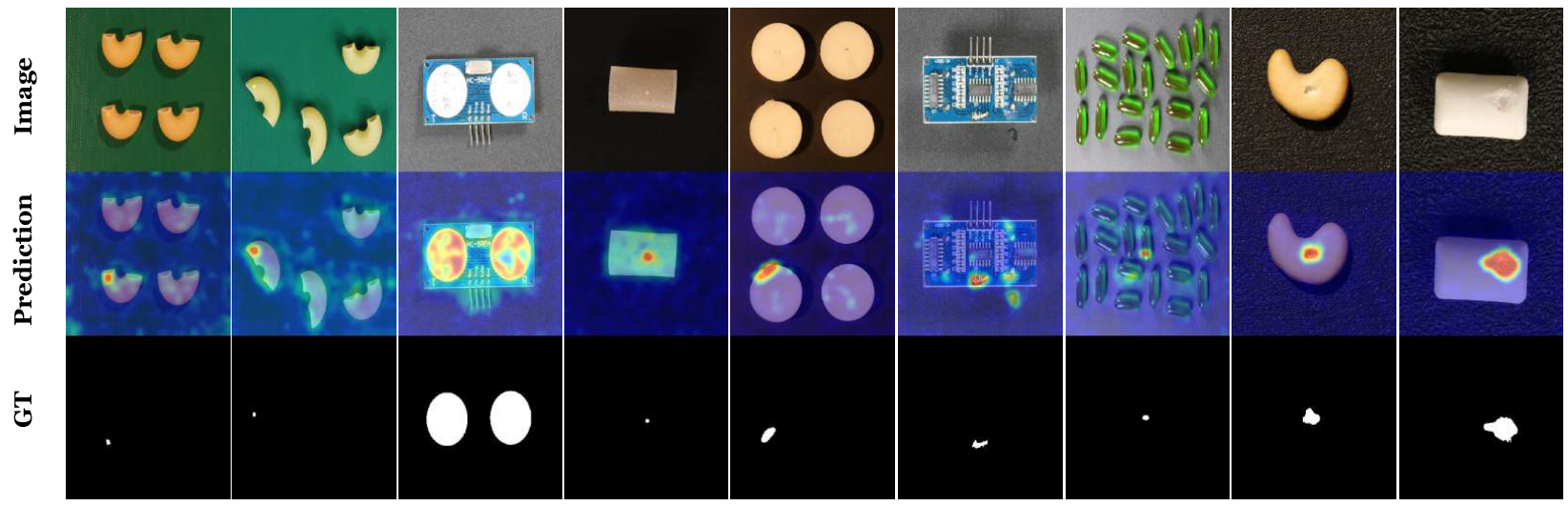}

\caption{
Qualitative anomaly localization results on VisA under the 2-shot setting.
Rows show input images, predicted anomaly maps, and ground-truth masks, respectively.
}
\label{fig:vis}
\end{figure*}

Table~\ref{tab:ablation_prompt_design} compares zero-shot transfer, normal-memory retrieval, and target prompt adaptation on VisA. All variants share the backbone and evaluation protocol, with one fixed random seed; variants using learned anchors share the same Stage-I checkpoint. The \emph{Stage-I only} baseline uses no target support images, memory bank, or parameter adaptation. All other variants use the same two normal images per target category, visual-memory construction, and score-fusion settings. The training duration is held fixed wherever target adaptation is enabled.

The Stage-I-only model achieves 83.9 image-level AUROC through zero-shot transfer. Adding a normal memory bank without updating model parameters raises this score to 87.6, showing the benefit of target normal references. With the same memory bank retained, the full model further improves I-AUC, I-AUPR, P-AUC, and PRO by 3.8, 3.4, 0.5, and 0.3 percentage points, respectively. This comparison isolates the additional gains from target prompt adaptation beyond normal-image retrieval. Separately, replacing the learned anchors with manual prompts while retaining the same target-adaptation procedure lowers image-level AUROC from 91.4 to 87.3. Together, these controls support combining auxiliary-learned semantic references with adaptation to target normal appearance.

To assess the effect of additionally updating the abnormal anchors, the \emph{constrained abnormal-anchor adaptation} variant retains the full normal-adaptation procedure and additionally updates the abnormal anchors via $\mathbf{a}_j=\operatorname{normalize}(\mathbf{a}_j^{(0)}+\boldsymbol{\delta})$. The shared residual is optimized to repel the anchors from normal global features and the 10\% of patches with the largest margin violations, using a margin of 0.05 relative to the fixed initial normal reference. Source-anchor preservation and $\|\boldsymbol{\delta}\|_2\leq0.2$ constrain this update. The additional losses update only $\boldsymbol{\delta}$, with the visual features and initial normal reference fixed. Keeping the abnormal anchors frozen improves all four metrics over this constrained alternative, supporting the proposed normal-side adaptation strategy.

On the normal side, removing the generic normal branch reduces image-level AUROC by 0.8 points and PRO by 0.4 points, indicating that inherited normal evidence complements the adapted class-specific prompts. Removing the low-rank residual or text-anchor regularization reduces image-level AUROC by 0.7 and 1.1 points, respectively, and also lowers the other three metrics. These results support combining generic and target-specific normal representations through residual adaptation with anchor regularization.

\subsection{Hyperparameter Sensitivity}
Tables~\ref{tab:sensitivity_tokens} and~\ref{tab:sensitivity_rank} vary the number of newly learnable tokens $J$ and the residual factorization width $\rho$ on VisA under the 2-shot setting, with all other settings fixed. Both parameters control the capacity of the class-specific normal prompts. Among the evaluated settings, $J=2$ and $\rho=8$ yield the highest image-level AUROC and AUPR and pixel-level AUROC, with tied-best PRO. Increasing either parameter further does not improve the scores.

\subsection{Visualization Results}
\label{subsec:visualization}

Figure~\ref{fig:vis} presents qualitative anomaly localization results on the VisA dataset under the 2-shot setting.
The visualization covers different product categories and defect patterns, including small localized defects, surface contamination, and structural abnormalities.
The predicted anomaly maps generally highlight regions that are consistent with the ground-truth masks, showing that the proposed method can localize abnormal areas using only a few normal samples from the target product.

\section{Conclusion}
\label{sec:conclusion}
We presented Anchor and Adapt, a two-stage prompt learning framework for few-shot industrial anomaly detection. Auxiliary anomaly supervision establishes transferable normal and abnormal anchors, while target normal supervision adapts product-specific normal appearance. This division reduces reliance on manually specified category-specific anomaly descriptions. Asymmetric adaptation preserves the learned anchors and updates an additional normal branch; the inherited and adapted normal representations jointly provide normal evidence under anchor regularization. Cross-dataset experiments on MVTec-AD and VisA, together with additional MPDD evaluation, demonstrate competitive detection and localization. Controlled ablations support the use of learned anchors and the proposed adaptation mechanisms, including gains beyond normal-memory retrieval.

\noindent

\bibliographystyle{IEEEtran}
\bibliography{references}

\end{document}